\documentclass[letterpaper, 10 pt, conference]{ieeeconf}  

\IEEEoverridecommandlockouts                              

\pdfoutput=1

\usepackage{graphics} 
\usepackage{epsfig} 
\usepackage{mathptmx} 
\usepackage{times} 
\usepackage{amsmath} 
\usepackage{amssymb}  
\usepackage{hyperref}
\usepackage{multirow}
\usepackage[table]{xcolor}
\usepackage{cite}

\title{\LARGE \bf
Read the Room: Adapting a Robot's Voice to Ambient and Social Contexts
}

\author{Paige Tutt\"os\'i$^{1}$, Emma Hughson$^{1}$, Akihiro Matsufuji$^{2}$, Chuxuan Zhang$^{1}$, and Angelica Lim$^{1}$
\thanks{*This work was supported by NSERC Discovery Grant 06908-2019. The authors thank P.J. Yazdian, M. Chamoux, S. Sanchez-Restrepo and Z.Y. Ou Yang for their valuable discussions, and the Rajan Family for their support.}
\thanks{$^{1}$P. Tutt\"os\'i, E. Hughson, C. Zhang, and A. Lim are with the School of Computing Science,
        Simon Fraser University, 8888 University Dr., Burnaby, Canada
        {\tt\small \{ehughson,ptuttosi,cza152,angelica\}@sfu.ca}}%
\thanks{$^{2}$A. Matsufuji is with Graduate School of System Design, Faculty of Computing Science,
        Tokyo Metropolitan University, 6-6 Hino city, Tokyo, Japan
        {\tt\small  matsufuji-akihiro@ed.tmu.ac.jp}}%
}

\begin{document}

\maketitle
\thispagestyle{empty}
\pagestyle{empty}

\begin{abstract}
How should a robot speak in a formal, quiet and dark, or a bright, lively and noisy environment? By designing robots to speak in a more social and ambient-appropriate manner we can improve perceived awareness and intelligence for these agents. We describe a process and results toward selecting robot voice styles for perceived social appropriateness and ambiance awareness. Understanding how humans adapt their voices in different acoustic settings can be challenging due to difficulties in voice capture in the wild. Our approach includes 3 steps: (a) Collecting and validating voice data interactions in virtual Zoom ambiances, (b) Exploration and clustering human vocal utterances to identify primary voice styles, and (c) Testing robot voice styles in recreated ambiances using projections, lighting and sound. We focus on food service scenarios as a proof-of-concept setting. We provide results using the Pepper robot's voice with different styles, towards robots that speak in a contextually appropriate and adaptive manner. Our results with N=120 participants provide evidence that the choice of voice style in different ambiances impacted a robot's perceived intelligence in several factors including: social appropriateness, comfort, awareness, human-likeness and competency.
\end{abstract}

\vspace*{-5mm}
\section{Introduction}
From the moment we wake up, to the moment we go to sleep, our day contains a variety of social situations in different contexts. For example, our first stop may be to the local bustling café, picking up our usual coffee to start the day. The end of the day may involve a date night at a fancy romantic restaurant, or a loud, rhythmic nightclub. In either case, these environments contain different lighting and music to set the mood, and people with whom we, hopefully, have the pleasure of interacting.

Humans have the unconscious ability to adapt their voice appropriately to different contexts and social situations. A significant portion of adaptive communication results from non-linguistic vocal features that can alter the meaning of phrases \cite{maruri21_interspeech}. A waiter may ask a customer, “Can I take your order?”, and this phrase may carry a different connotation depending on whether the waiter asks the phrase in a casual cafe during a rush, or an upscale bar with highbrow clientele. Nonetheless, the ambient surrounding plays a role in how one changes their voice to convey meaning to others, and being able to reproduce and interpret these non-linguistic features plays a significant role in human social intelligence.

\begin{figure}[]
      \centering
      \includegraphics[scale=0.33]{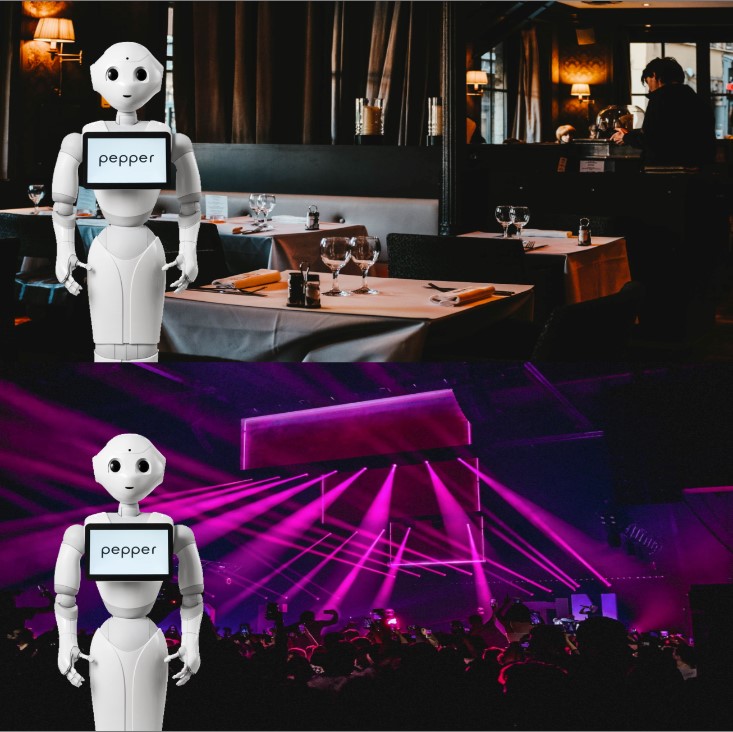}
      \caption{Robots may be deployed in varied ambiances, from cozy formal dining to loud nightclubs. How should their voices change?}
      \vspace*{-6mm}
      \label{diagram}
\end{figure}


 The need for a robot to adapt itself appropriately in ambient and social environments proves important when integrating robots into humans’ everyday lives \cite{context_and_robot_voice,progress_on_robotics_lit_review,what_makes_a_robot_social}. We hypothesize that correct adaptation can improve both the robot's perceived awareness and intelligence. Yet, the exact way to adapt robotic voices has not been thoroughly explored due in part to the difficulty in collecting clean recordings of realistic data in noisy and crowded environments, i.e. to separate  voice and background noise. As such, source voice data for text-to-speech systems (TTS) is typically recorded and tested in quiet office environments. The current study utilizes a readily available video-conferencing platform equipped with ambient sounds to aid actors in adapting their voice to fit the target environment. In addition to this novel data collection protocol, a comprehensive user perception study is conducted using the Pepper robot\footnote{https://www.softbankrobotics.com}. Altogether, the current study aims to bridge the aforementioned gaps in the literature by building upon previous work to:
 \vspace*{-1mm}
\begin{enumerate}
\item implement a novel protocol for collecting realistic contextual audio voice data.
\item assign robot voices to different ambient contexts by investigating and clustering human vocal features.
\item assess human perception of robot voices to better understand what is considered appropriate in different ambient settings.
\end{enumerate}
 \vspace*{-2mm}
 
\section{Related Work}
\subsection{Human Contextual Voice Modifications}
Human vocal modifications are commonly used to create ‘deliberately clear speech,’ when a listener is experiencing reduced comprehension \cite{confluent_talker_and_listener}. These modifications may occur when the environment is causing auditory hindrance, as is the case with distant speakers \cite{vocal_effort_acoustic_environments}, distorted transmissions \cite{acoustic_phonetic_adverse_listening_conditions}, or noisy spaces \cite{acoustic_phonetic_adverse_listening_conditions}. One of the most well researched vocal phenomena is the Lombard effect \cite{lombard_origin}, an involuntary increase in vocal effort, often due to the presence of background noise \cite{acoustic_phonetic_adverse_listening_conditions}.  Modifications are also often listener-specific, as is the case in infant, child \cite{whats_new_pussycat}, hearing impaired \cite{mommy_speak_clearly}, and machine directed speech \cite{prosodic_and_intelligibility}.  In some cases, speech is not modified for clarity, but rather to communicate a specific emotion or purpose, such as politeness~\cite{polite}. 

\subsection{Current State of Robot Contextual Voices}

Several studies have shown that humans show preferences for voice depending on context and task \cite{context_and_robot_voice, perceptual_effect_ros, providing_route_directions, adaptive_speech_cog_robots}. In \cite{context_and_robot_voice}, participants rated the appropriateness of different robot voices given varied contexts including schools, restaurants, homes and hospitals. They found that even given the same physical appearance, participants selected varying voices depending on context and concluded that a robot voice created for a specific context is likely not generalizable. Studies suggest the incorporation of context-based methods such as sociophonetic inspired design \cite{voice_as_a_design_material} and acoustic-prosodic adaption to match user pitch \cite{voice_adaptation_robot_learning_companion} or incremental adaptation of loudness to the user's distance \cite{HRI_incremental_speech_adaptation}. 

To the best of our knowledge, only one study has explored adaptation of a single robot's voice to its acoustic environment, by adjusting volume only based on environmental noise levels \cite{volume_adaptation_telepresence_system}. Overall the literature has focused on adaptation of loudness in ambient environments. We expand on the literature by applying knowledge from studies on human voices, where we understand that adaptive appropriateness relies on more than loudness features alone, to address the observed need for more adaptive robot voices. 

\section{Understanding Human Ambiance Adaptation}
How do humans change their voices depending on the ambiance? We first describe our method for ambiance-adapted voice data collection, validation and analysis.

\subsection{Zoom Data Collection Protocol}\label{zoom}

We propose a method of virtual data collection using readily available tools
that will allow researchers to collect data with no physical human interaction. Although collecting data in an actual restaurant is preferred, it is difficult to collect in-the-wild audio that is high quality and free of the ambient background noise. Moreover, targeted ambient environments are often resource intensive to obtain or control. As such, one of the novelties of the current study was how we overcame the above hurdles by devising a protocol that mimicked naturalistic ambient environments over Zoom\footnote{www.zoom.us}.

Zoom is a teleconferencing program that allows individuals to communicate from anywhere in the world. It also allows for the use of virtual backgrounds and sharing of sound. In this study, we asked pairs of English speakers to listen to ambient sounds while conversing with one another in the roles of a waiter and a restaurant-goer using their personal laptop, headphones and microphone. There were a total of 6 ambient sounds and one additional baseline measure that included no sound and a black background. 

In addition to sound, the speakers were asked to change their Zoom background to an image that was pre-selected to match the given ambiance. A brief description of the ambiance (e.g. food available, brightness level, business of location, etc.) and character roles (waiter or restaurant-goer) were provided at the beginning of each ambient condition to induce vivid imagery. Between each ambient condition there was a one minute period to update participants' Zoom backgrounds and prepare for the next condition. This served as a washout to reduce carry-over effect from the previous condition. The ambient contexts included: fine dining, café, lively restaurant, quiet bar, noisy bar, and nightclub. The ambient sounds can be listened to \href{https://rosielab.github.io/ambiance}{here}\footnote{https://rosielab.github.io/ambiance}.

\begin{figure}[]
      \centering
      \includegraphics[scale=0.37]{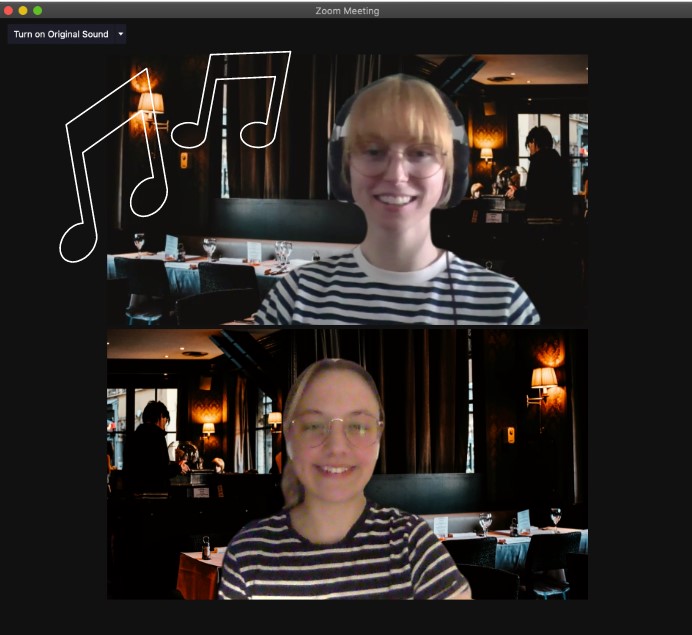}
      \caption{Zoom virtual backgrounds and ambient sounds through headphones were used for data collection.}
      \vspace*{-5mm}
      \label{zoomexp}
\end{figure}

The speakers first read a brief description of their character at the restaurant. They then read from a script that was tailored for the given ambiance, i.e., food and drink choices matched what is usually offered at the given restaurant. Consistency amongst scripts allowed for comparison of speech features across each condition. The subtle differences that did occur between scripts was solely to create a more realistic environment and reduced redundancy to maintain participant attention. The experiment was repeated in an unscripted manner, however, the conversations appeared uncomfortable and unnatural and were not included in the analysis.


The dataset consisted of 8 females undergraduate students (age not collected) with experience in improvisation, theatre or customer service. We used only female voices (biological sex) in order to match the source voice of Pepper’s TTS, which is from a female voice actor. Although the dataset is small, this is often the case in psycho-acoustic studies \cite{ARIAS2018R782}; we focus instead on quality of the data through our validation study (see Sec. \ref{val_study}). Altogether, we collected 837 utterances, resulting in 685 utterances post-validation. 

\subsection{Voice Validation Study}\label{val_study}
The voices were validated using a battery of 7-point Likert scales (1: strongly disagree, 7: strongly agree) to assess the speaker's ability to adapt to the given ambiance as well as the effectiveness of the virtual ambiance in inducing adaptation.  A voice ``passage" is single one-sided conversation of an actor in a given ambiance. Each passage was validated by participants proficient in English (N=12, mean age group = 25-35, 4 males, 8 females). Participants listened to each of the passages overlaid on either: (1) the ambiance the passage was recorded in, or (2) a mismatched ambiance. The validation questions loaded onto 4 primary factors: 1. Social appropriateness, 2. Ambient awareness, 3. Comfort, 4. Clarity. The full battery can be found in the Appendix.

 Questions on clarity were taken from the well-used MOS. To the best of the authors' knowledge, no validated measures exist clearly targeting the other factors, especially those that can be used out of context of a larger battery. As we were primarily interested in social and ambiance adaption, we removed a passage if it received a majority vote of less than 4 (neutral) in any question loading onto social appropriateness and ambiance awareness, meaning that the majority of validators disagreed that this passage was socially appropriate or ambiance aware.
 \begin{figure}[]
      \centering
      \includegraphics[scale=0.63]{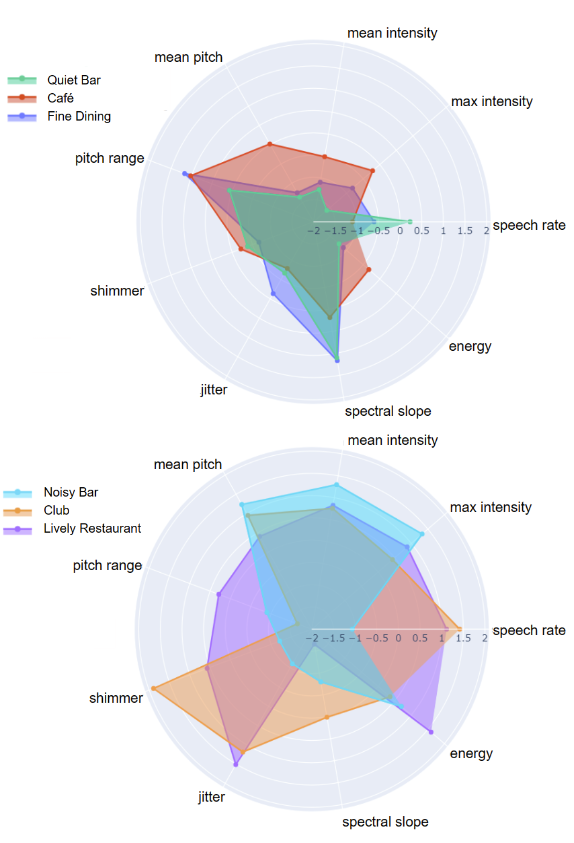}
      \caption{Audio features across speakers as a difference from the baseline.}
      \vspace*{-7mm}
      \label{radar}
\end{figure}

\subsection{Voice Analysis and Feature Extraction}

In order to observe how humans modify their voice, we collected 10 features, which can be grouped by (1) loudness, (2) spectral, including pitch, and (3) rate-of-speech. Our toolbox for vocal feature extraction can be found \href{https://github.com/ehughson/voice_toolbox}{here}\footnote{https://github.com/ehughson/voice\_toolbox}.

\subsubsection{Loudness Features}
 Increase of vocal intensity, often leading to Lombard speech, is commonly employed in noisy environments \cite{acoustic_phonetic_features_analysis}. As such, we collected 3 loudness features: (a) mean intensity, (b) energy, (c) maximum intensity. Mean intensity and energy features were calculated using the Praat \footnote{http://www.praat.org/} library via Parselmouth\footnote{https://parselmouth.readthedocs.io/en/stable/}. Librosa \footnote{https://librosa.org/doc/main/index.html} was used to calculate maximum intensity (power) following the formula provided in Section 1.3.3 of \cite{fundamentals_of_music}. 

\subsubsection{Spectral Features}
We collected 5 spectral features: (a) median pitch, (b) pitch range, (c) shimmer, (d) jitter, and (e) spectral slope. Parselmouth was used to extract (a)-(d). Median pitch and pitch range (the difference between the minimum and maximum pitch in a given segment) are calculated in Hz. Local shimmer and local jitter, variations in the fundamental frequency, are perceived as vocal fry and hoarseness, respectively \cite{jitter}. Spectral slope gives an indication of the slope of the harmonic spectra. For example, -12dB slope may indicate a falsetto voice, and -3dB slope can indicate richer vocal tones \cite{spectralslope}. Spectral slope was calculated using Parselmouth and Librosa with the formula provided in Section 3.3.6 of \cite{audioprocessing}.

\subsubsection{Rate-of-Speech Features}

We collected 1 rate-of-speech feature: syllables per second. Syllables per second was the number of syllables over the duration extracted using Praat scripts \footnote{https://github.com/drfeinberg/PraatScripts/blob/master/syllable\_nuclei.py}.

\subsubsection{Trends in Voice Data}
Radar plots constructed from our collected dataset are displayed in Fig. \ref{radar}. We can observe that although voices in the quiet ambiances appear to share vocal features, those in each of the loud ambiances appear quite different. The average voice in the bright, high arousal lively restaurant with fast music (140 BPM) showed a higher pitch range and energy with a low spectral slope when compared to the other ambiances. The average nightclub voice, by comparison, showed high shimmer, jitter and spectral slope, with a high pitch and low pitch range, suggesting a voice consistent with Lombard speech. Lastly, voices in the loud bar showed moderate values for most features aside from high loudness. First, these trends suggest that adapted voices indeed vary by features other than loudness. As loudness is already well represented in the literature, continuing forward we will freeze loudness features across voice styles to focus on the contribution of other vocal features. Second, we observed that the quiet voices may be a single style, and as such continue forward with clustering.

\section{Robot Voice Styles Using Clustering}
As the development of a robot voice is resource-intensive, we aimed to find a minimal set of voice styles needed to work well in our ambiances. We used scikit-learn \footnote{https://scikit-learn.org/stable/} to scale and cluster our voice data. First, K-means clustering was performed on voice features collected in Section \ref{zoom}. Robust scalar was used to normalize each feature. Once scaled, the number of clusters, set from 2 to 10, were tested and a silhouette score was collected to determine the appropriate amount of clusters. Although silhouette score was lower for a larger number of clusters, when observing the cluster composition, we observed that more often than not the number of utterances for each ambiance was equal across clusters. This suggests vocal features for these clusters differ on an utterance level rather than ambiance level. Exploring utterance level information, such as positive or negative phrases is left as future work. As such, we determined three clusters was appropriate, and that three voice styles would need to be selected or designed.

\subsection{Human Voice Cluster Analysis}
\begin{figure}[]
\hspace*{-0.8cm}
      \centering
      \includegraphics[scale=0.45]{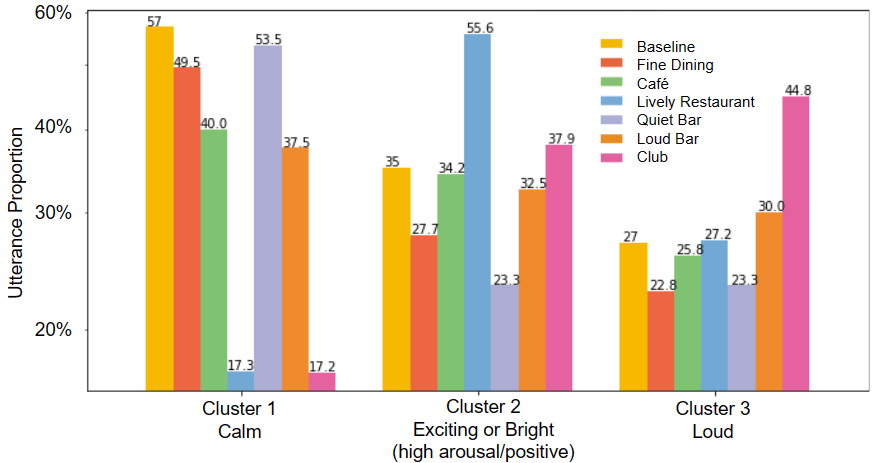}
      \caption{Which voice styles are associated with which ambiance? We visualize the proportion of each ambiance's utterances represented by each voice style.}
      \label{barplot}
      \vspace*{-8mm}
\end{figure}

\begin{figure}
      \centering
      \includegraphics[scale=0.48]{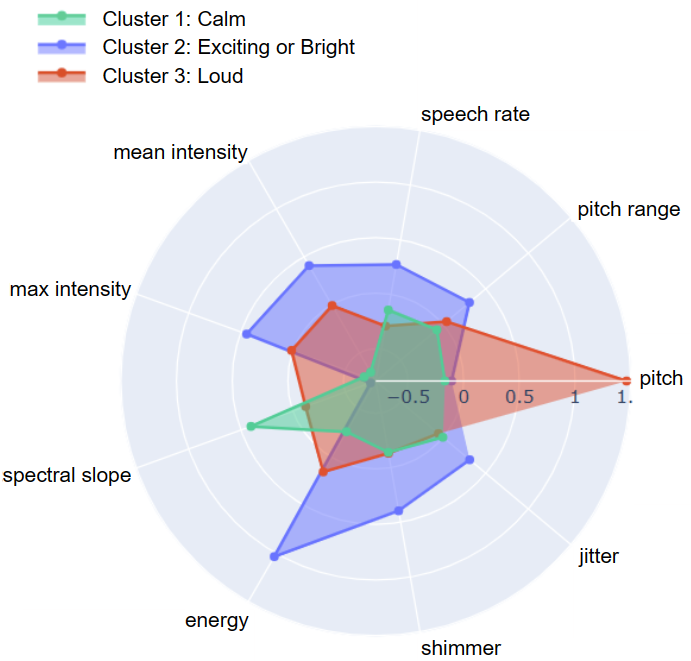}
      \caption{What do the voice styles sound like? Feature analysis of the three voice styles derived from human voice cluster centers.}
      \label{radar3}
      \vspace*{-8mm}
\end{figure}

In which ambiances are the voice styles (derived from clusters) primarily used, and what do they sound like? Towards answering these questions, we created two types of plots: (1) for each cluster, the proportion of each ambiance's utterances using this voice style (Fig. \ref{barplot}), and (2) a radar plot of each cluster's centres to depict the voice style's characteristics (Fig. \ref{radar3}). In Fig. \ref{barplot},  we see that utterances belonging to the first cluster occur mostly in the fine dining and quiet bar ambiances, and do not occur in our two highest arousal ambiances. We therefore designate it as ``calm". The perception rating of each ambiance can be seen in our perception study Fig. \ref{heatmap2}. Utterances in the second cluster occur most often in the lively restaurant, a high arousal, positive ambiance as well as the brightest. We designate this second cluster as ``exciting or bright". The third cluster is occurring primarily in the loudest ambiance, the nightclub. We suspect this is a Lombard voice and designate the name ``loud". One ambient context of note is the loud bar, which appears to occur almost equally across clusters. This may have occurred for several reason including: a split in adaptation between the server and the customer, different interpretations of the atmosphere by the speakers, or strong utterance level differences for this ambient context.

Looking at the radar plot of the features for each cluster's centers (Fig. \ref{radar3}), we see that the ``exciting or bright" cluster has high energy, speech rate, pitch range, jitter and shimmer. This appears to fit the assessment of this being a happy and excited voice.  For the calm cluster, we see moderate values on most spectral and pitch features with the highest spectral slope, which can indicate a richness in the voice \cite{spectralslope}. Lastly, looking to the loud cluster, it does indeed appear to be Lombard: we see the low speech rate, lower pitch range and high pitch that is quintessential of this voice style. Lastly, we see a parallel from the initial analysis of the human voices (Fig. \ref{radar}), where the quiet voices belong to a single cluster, yet voices in the loud ambient contexts differ based on a number of factors. Our next step is to select robot voices in these styles, towards a robot voice perception study.

\label{HUMANCLUST}

\begin{table}[]
 \vspace{1mm}
 \caption{Comparison of features for voice clusters and Pepper voices.}
 \vspace{-2mm}
 \label{table1}
\begin{tabular}{|p{1.2cm}|p{1.5cm}|p{1.9cm}|p{2.2cm}|}
 \hline
\multicolumn{4}{|c|}{Human Voice Clusters} \\
\hline
  & Pitch(Hz) & Pitch Range(Hz) & Speech Rate(syll/s)\\
 \hline
Calm & Medium & Medium & Medium\\
Exc/bright & Medium & High & High \\
Loud &   V. High  & Medium & Low\\
\hline
\multicolumn{4}{|c|}{Pepper Voices} \\
\hline
 Neutral & Low(350) & Medium(575) & Medium(4.18)\\
Joyful & V. High(408) & V. High(699) & High(4.41)\\
 Didactic & Medium(367) & Low(558) & Low(3.71)\\
``Lombard" & V. High(435) & Low(467) & Low(3.71) \\
 \hline
\end{tabular}

{\parbox{3.4in}{
\vspace{1mm}
\footnotesize{Note: For human voices the values were assigned based on the value of the cluster center following robust scaling. Values between -0.25 and 0.25 were assigned medium, those over 0.25 high and below -0.25 low. For Pepper's voices, features were scaled separately from the human features using MaxAbsScaler with the values in the table.}
}}
\label{overallstats}
\vspace*{-7mm}
\end{table}

\begin{table}[]
 \vspace{1mm}
 \caption{Comparison of features for voice clusters and Pepper voices.}
 \vspace{-2mm}
 \label{table1}
\begin{tabular}{|p{1.2cm}|p{1.5cm}|p{1.9cm}|p{2.2cm}|}
 \hline
\multicolumn{4}{|c|}{Voice Clusters} \\
\hline
  & Pitch(Hz) & Pitch Range(Hz) & Speech Rate(syll/s)\\
 \hline
Calm & Medium & Medium & Medium\\
Exc/bright & Medium & High & High \\
Loud &   V. High  & Medium & Low\\
\hline
\multicolumn{4}{|c|}{Pepper Voices} \\
\hline
 Neutral & Low(350) & Medium(575) & Medium(4.18)\\
Joyful & V. High(408) & V. High(699) & High(4.41)\\
 Didactic & Medium(367) & Low(558) & Low(3.71)\\
``Lombard" & V. High(435) & Low(467) & Low(3.71) \\
 \hline
\end{tabular}

\vspace*{-5mm}
\end{table}

\subsubsection{Voice Clusters to Robot Voice Styles}
We extracted voice features, once again using our voice toolbox, from Pepper's three available voice styles: Neutral, Joyful and Didactic. Given the human clusters, our next step is to map our voice styles to our robot's voice. We aim to make as little changes to Pepper's voices as possible to demonstrate that, although these voices have been expertly curated to match this robot, they may not be applicable in all situations and care needs to go into understanding in which context each voice should be applied. We also leave synthesizing new voices as future work, as the proposed data-driven voice selection method does not need access to the robot's original voice actor, nor a large amount of training data and is therefore more applicable to researchers working in low resource environments. Moreover, it is important to maintain the identity and overall features that make these voices appropriate for this robot.

A comparison of the cluster features with the features of each of Pepper's voices can be found in Table \ref{table1}. We only include those features that can be modified through Pepper's TTS mark-up language. Again, we chose not to include loudness features as we aim to maintain Pepper's voice at a constant volume to explore how the alteration of non-loudness features can effect perception of the voice. Based on the results in the table we decided to equate the ``exciting or bright" cluster with Pepper's Joyful voice and the ``calm" cluster with Pepper's Neutral voice. For the ``loud" cluster, we did not have a particularly close match given the high pitch and low pitch range. We noted that Pepper's Didactic voice had the lowest pitch range and speech rate, therefore, we decided to build our ``Lombard" voice with this Didactic base increasing the pitch by 130\%.


\section{Perception Study}

To better understand human perception of our selected ambient robot voices we conducted a perception study. 
We aimed to address three research questions (RQs):\\
\textbf{RQ 1}: What voice styles are preferred for the target ambiances?\\
\textbf{RQ 2}: What ambiance-specific voice styles can increase perceived appropriateness and user comfort?\\
\textbf{RQ 3}: Does improved appropriateness and comfort occur with an increased perception of competency, awareness and human-likeness, i.e. overall intelligence?

We recruited 120 students to participate: 71 men, 44 women, 4 non-binary and 2 who preferred not to answer. The participants were primarily between the ages of 18-24 (N=114). Participants were required to have normal hearing, be over the age of 18 and have a proficient grasp of English in order to participate. The students were primarily recruited through an introductory computing science course. This study received research ethics approval.

\begin{figure}[]
      \centering
      \includegraphics[scale=0.25]{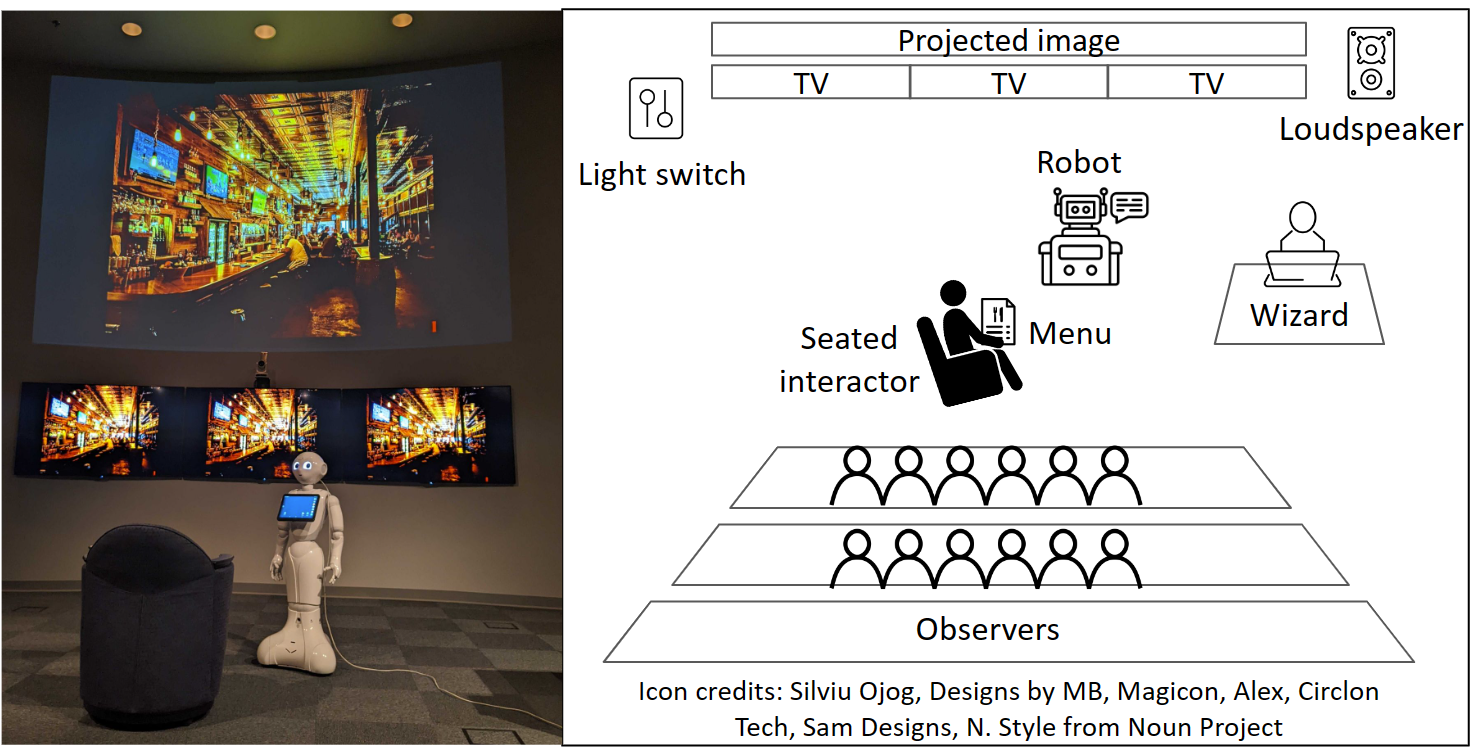}
      \caption{Experimental setup for user perception study, and the ``noisy bar" condition (left).}
      \label{diagram}
      \vspace*{-7mm}
\end{figure}

We ran our study in a presentation theatre within a university campus as a controlled proxy for actual restaurants. In order to replicate our desired ambiance, we projected the same restaurant image used to collect voice data on the wall, as well as three TV screens at the front of the hall. Additionally, we used the ambient speakers available within the room to play the music used in the data collection. Fine dining, café and quiet bar were set to 50 dB, lively restaurant and noisy bar were set to 55 dB, and nightclub was set to 60 dB using a decibel meter. Lastly, we controlled the lighting by keeping the blinds closed and turning on and off lights to replicate `dark', `warm and dim', or `bright' environments. 
The ambiances were presented in random order for each trial. 

The participants interacted one at a time, for a single interaction with the Pepper robot while the others observed. For each ambiance, three interactions took place using the  script below, each with a different voice style. Pepper was controlled using Wizard of Oz, with Pepper's ALTextToSpeech module. Pepper said, \emph{``Hello, I hope you are doing well. I hope you have had a chance to look at the menu. I recommend the daily special. Anyways, what can I get you?"} The interacting participant (hereafter, ``interactor") then verbally replied, choosing an item from the paper menu in front of them.  Pepper then replied, \emph{``Sure, I can definitely do that. I will be right back with your order."} During the conversations, Pepper's tablet displayed a black screen and no gestures were used. Basic Awareness and idle body motions were activated to allow the robot to appear lifelike while also allowing the participants to focus on the voice.

Once the conversation was completed, all participants, both the interactor and the observers, completed an online questionnaire. During this time, a black screen was shown and no music was played. The questions began by asking whether you were the one who interacted with the robot. The remaining questions used a 7 point Likert scale (1: strongly disagree, 7: strongly agree) as follows:

\begin{itemize}
 \footnotesize
\item Pepper's voice is socially appropriate for the scene
\item Pepper is aware of the surrounding ambiance
\item Pepper makes me feel comfortable
\item Pepper sounds human-like
\item Pepper is a competent server
\end{itemize}

To the best of the authors' knowledge no validated measures exist clearly targeting all of these factors, specifically social appropriateness within in a scene and ambiance awareness; especially without the inclusion of numerous extraneous questions. At the end of each ambiance the participants answered a ``scene change" question set while the next ambiance was set up. These questions included which of the three voices they would prefer as their server, and questions to assess the accuracy of the ambiance. First, we asked whether ``The ambiance matched the prompt" with a 100 point slider from disagree to agree. The subsequent questions were on a sliding scale from 1 to 100 with the prompt ``The ambiance is:" with slider scales of Dark-Bright, Quiet-Noisy, Casual-Formal, Calm-Exciting, Cold-Warm, Negative-Positive, Enclosed area-Open area. For each trial, we completed 18 of the previously described short conversations. The three voices from Sec. IV-B, Neutral, Joyful, and ``Lombard" were used in each of the 6 different ambiances: a fine dining restaurant, a café, a lively restaurant, a quiet bar, a loud bar and a nightclub. 

\section{Results}
We first tested all results for significant differences between the ratings of the observer and the interactor and none were found; therefore, moving forward, observers and interactors will be treated as a part of the same group.
\subsection{Ambiance Analysis}
\begin{figure}[t]
      \centering
      \includegraphics[scale=0.28]{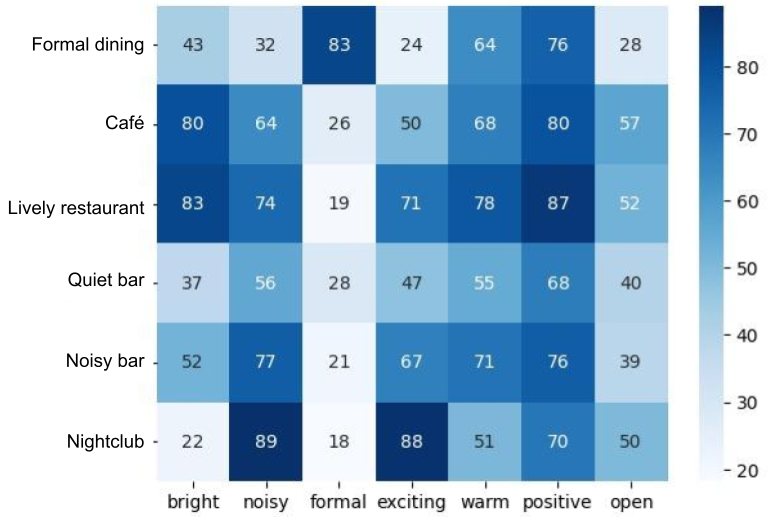}
      \caption{Perception of ambient characteristics (rated from 1-100) for each condition based on participant results.}
      \label{heatmap2}
      \vspace*{-5mm}
\end{figure}

We found that in general, our physical ambiances were a good match to the given prompt (N=714, $\mu$=81.25, $\sigma$=20.10) when rated on a scale from 1 to 100. The agreement was lowest for the quiet bar (N=120, $\mu$=77.34, $\sigma$=23.56) and highest for the fine dining restaurant (N=120, $\mu$=85.63, $\sigma$=17.86). The average ratings for each ambiance feature on a scale from 1-100 can be seen in Fig. \ref{heatmap2}.

\subsection{RQ 1: Preferred Voices for Ambiances}
\begin{figure}[t]
      \centering
      \includegraphics[scale=0.48]{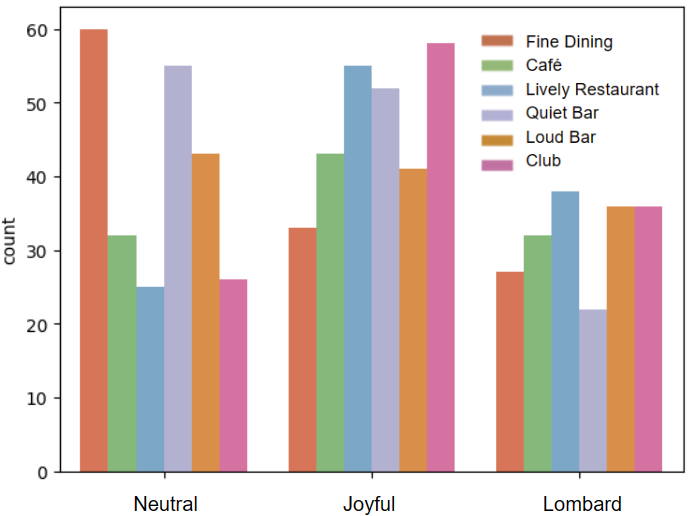}
      \caption{Number of participants choosing each voice as first preference for each ambiance. Significance levels reported in Table III.}
      \label{1st}
      \vspace*{-6mm}
\end{figure}

We completed Chi Squared Goodness of Fit tests, followed by confidence intervals (CIs) to assess preference for specific voices in a given ambiance. At $\alpha$ = 0.001 we found that a Neutral voice was least often selected as the preferred server in a lively restaurant, and a nightclub, and that the ``Lombard" voice was least often selected as the preferred sever in the fine dining and quiet bar. Moreover, the Joyful voice was least often selected as the last choice in the lively restaurant. So although Joyful showed no significance as the preferred voice, we know that it is not the least preferred. Full significant results can be seen in Table \ref{chi}, and a plot of first choice voices can be seen in Fig. \ref{1st}.

\begin{table}[t]
 \hspace*{-2mm}
 \caption{Chi-Square Preference Results}
 \label{chi}
 \vspace{-3mm}
\begin{tabular}{ |p{1.3cm}|p{0.74cm}|p{0.59cm}|p{0.7cm}|p{0.55cm}| p{0.75cm}| p{0.9cm}| }
 \hline
\multirow{2}{*}{Ambiance} &
  \multicolumn{3}{c|}{Count} &
  \multirow{2}{*}{$\chi^{2\;2}$} &
  \multirow{2}{*}{P-value} &
  \multirow{2}{*}{CI$^3$}\\
  
   & Neutral & Joyful & 'Lomb.' &  & &    \\
\hline
 \multicolumn{7}{|c|}{\MakeUppercase{First Choice}} \\
 \hline
 Fine Dining   & 60 & 33 &  \textbf{27} & 15.45 & 0.001** &  13-32\% \\
 Lively Res.   & \textbf{25} & 55 & 38 & 11.51 & 0.003** & 12-31\% \\
 Quiet Bar  & 52 & 49 & \textbf{19} & 15.49 & 0.001** & 5-22\% \\
 Nightclub  & \textbf{26} & 58 & 36 & 15.49 & 0.001** & 12-31\%\\
 \hline
  \multicolumn{7}{|c|}{\MakeUppercase{Last Choice}} \\
 \hline
  Lively Res.   & 58 & \textbf{25} & 35 & 14.65 & 0.001** & 12-31\% \\
 \hline
 
\end{tabular}
{\parbox{3.4in}{
\footnotesize 
1. \textbf{Bold} values highlight the voices selected at a significantly different ratio.\\
2. DF = 2, N = 120 (118 for Lively Restaurant)\\
3. Confidence interval on selection proportion is reported only for the voice that was selected at a significantly different ratio.\\
4. Due to space limitations we only report significant values.
}
}
\vspace*{-7mm}
\end{table}

\subsection{RQ 2: Social Appropriateness and Comfort}
\begin{figure}[]
      \centering
      \includegraphics[scale=0.23]{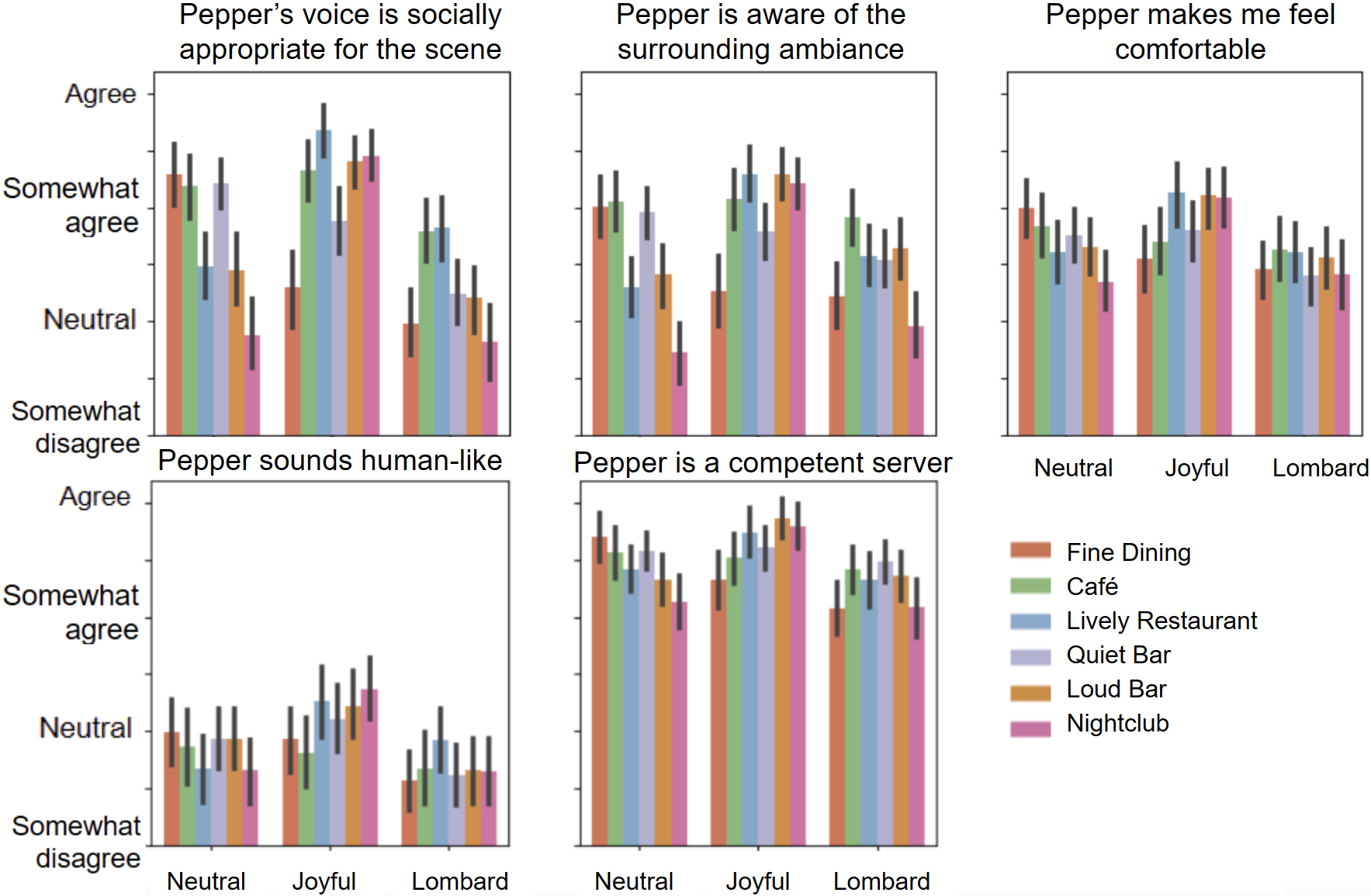}
      \caption{Likert score results for all questions resulting in at least one significantly different result across voices (full details in Table \ref{anova}).}
      \label{likert}
      \vspace*{-3mm}
\end{figure}

For both RQ1 and RQ2 we used a one-way ANOVA followed by a post-hoc Tukey HSD and set $\alpha$=0.001 to assess improvements in Likert ratings between voices, within each ambiance, for our targeted questions. A full table of these results can be found in Table \ref{anova} and a visualization of the Likert scores in Fig. \ref{likert}.

We found that a Neutral voice was more socially appropriate for fine dining and that a Joyful voice was more socially appropriate in a lively restaurant, noisy bar and nightclub. In addition, a Joyful voice made participants feel more comfortable in a nightclub.

\begin{table*}[t]
 \hspace*{-2mm}
 \caption{Results for Difference of Mean Statistical Tests}
 \vspace{-2mm}
 \label{anova}
\begin{tabular}{ |p{2.0cm}|p{0.75cm}|p{0.75cm}|p{0.75cm}|p{0.75cm}|p{0.75cm}|p{0.75cm}| p{0.55cm}|p{0.75cm}|p{0.8cm}|p{2.0cm}|p{1.5cm}| }
 \hline
\multirow{2}{*}{Ambiance} &
  \multicolumn{2}{c|}{Neutral} &
  \multicolumn{2}{|c|}{Joyful} &
  \multicolumn{2}{|c|}{'Lombard'} &
  \multirow{2}{*}{DF} &
  \multirow{2}{*}{N} &
  \multirow{2}{*}{F} &
  \multicolumn{2}{c|}{P-value}\\
  
   & $\mu$ & $\sigma$ & $\mu$ & $\sigma$ & $\mu$ & $\sigma$ & & & & One-Way Anova & Tukey HSD\\
    \hline
 \multicolumn{12}{|c|}{\MakeUppercase{Socially Appropriate}} \\
 \hline
 Fine Dining   & \textbf{5.30} & \textbf{1.49} & 4.30 & 1.73 &  3.97 & 1.66 & 2 & 120 & 22.58 & $<0.001$*** & $<0.001$***\\
 Lively Restaurant   &  4.48 & 1.63 & \textbf{5.68} & \textbf{1.14} & 4.82 & 1.59 & 2 & 118 & 18.93 & $<0.001$*** & $<0.001$***\\
 Noisy Bar   & 4.41 & 1.54 & \textbf{5.29} & \textbf{1.17} & 4.61 & 1.47 & 2 & 120 & 20.07 & $<0.001$*** & $<0.001$***\\
 Nightclub   & 3.88 & 1.54 & \textbf{5.46} & \textbf{1.26} & 3.28 & 1.87 & 2 & 120 & 38.28 & $<0.001$*** & $<0.001$***\\
 \hline
 \multicolumn{12}{|c|}{\MakeUppercase{Ambiance Aware}} \\
 \hline
  Fine Dining   &  \textbf{5.01} & \textbf{1.54} & 4.26 & 1.65 & 4.22 & 1.63 & 2 & 120 & 9.43 & $<0.001$*** & $<0.001$***\\
  Lively Restaurant   &  4.31 & 1.42 & \textbf{5.29} & \textbf{1.26} & 4.58 & 1.58 & 2 & 118 & 13.92 & $<0.001$*** & $<0.001$***\\
  Noisy Bar   & 4.41 & 1.42 & \textbf{5.29} & \textbf{1.17} & 4.64 & 1.48 & 2 & 120 & 12.46 & $<0.001$*** & $0.001$**\\
 Nightclub   & 3.73 & 1.59 & \textbf{5.22} & \textbf{1.27} & 3.96 & 1.54 & 2 & 120 & 35.25 & $<0.001$*** & $<0.001$***\\
 \hline
  \multicolumn{12}{|c|}{\MakeUppercase{Comfort}} \\
 \hline
  Nightclub   &  4.35 & 1.45 & \textbf{5.09} & \textbf{1.40} & 4.41 & 1.56 & 2 & 120 & 9.43 & $<0.001$*** & $<0.001$***\\
 \hline
   \multicolumn{12}{|c|}{\MakeUppercase{Human-likeness}} \\
 \hline
  Nightclub   & 3.66 & 1.57 & \textbf{4.37} & \textbf{1.59} & 3.65 & 1.60 & 2 & 120 & 8.05 & $<0.001$*** & $0.002$**\\
 \hline
  \multicolumn{12}{|c|}{\MakeUppercase{Competency}} \\
 \hline
  Noisy Bar   &  5.32 & 1.43 & \textbf{5.87} & \textbf{1.35}  & 5.36 & 1.45 & 2 & 120 & 7.91 & $<0.001$*** & $0.002$**\\
 Nightclub   &  5.13 & 1.37 & \textbf{5.80} & \textbf{1.06}  & 5.08 & 1.43 & 2 & 120 & 11.46 & $<0.001$*** & $<0.001$***\\
 \hline
 
\end{tabular}
{\parbox{6.8in}{
\footnotesize 
1. \textbf{Bold} values highlight the preferred voice.\\
2. Due to space limitations we only report those voices that were found to have a significantly different Likert rating than both other voices. In all cases the post-hoc P-value was equal for all pairings with the preferred voice.
}
}
\vspace*{-4mm}
\end{table*}

\vspace*{-3mm}
\subsection{RQ 3: Awareness, Competency, and Human-likeness}
We found that a Neutral voice was perceived as more ambiance-aware for fine dining and that a Joyful voice was perceived as more ambiance-aware in a lively restaurant, noisy bar and nightclub. In addition, the robot with a Joyful voice was perceived as more human-like and competent in a nightclub and more competent in a noisy bar.

\begin{figure}[]
      \centering

      \includegraphics[scale=0.6]{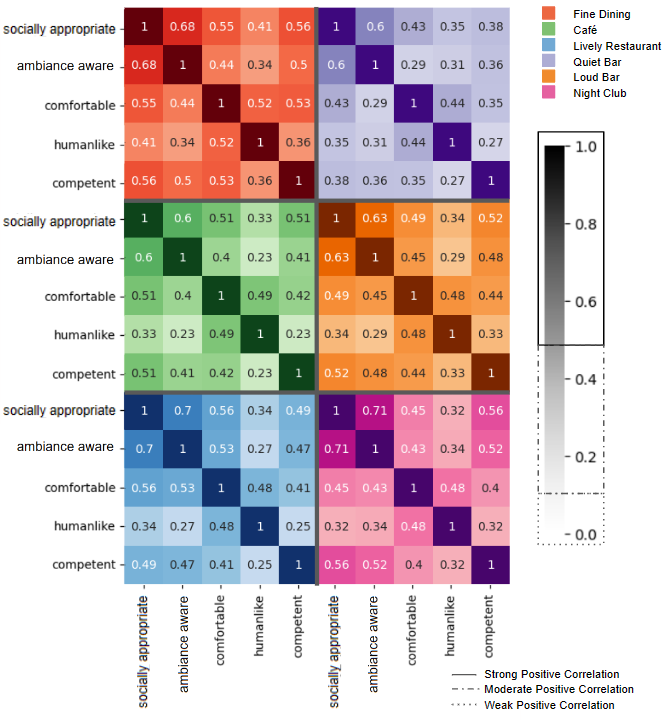}
      \caption{Pearson's R correlations between perception study responses for each ambiance.}
      \label{corr}
      \vspace*{-7mm}
\end{figure}

Correlation results can be seen in Fig. \ref{corr}. We find the strongest positive correlations between social appropriateness and ambiance awareness in all ambient contexts. Strong positive correlations are seen between social appropriateness, competency, and comfort for fine dining and the café, with competency in the nightclub and loud bar, and with comfort in the lively restaurant. Comfort shows strong positive correlations with competency and human-likeness for fine dining and ambiance awareness in the lively restaurant. Lastly, ambiance awareness shows strong positive correlations with competency for fine dining and the nightclub, and comfort in the lively restaurant. All other correlations are moderately positive, which overall tells us that by increasing social and ambiance appropriateness we can improve perceptions of social and overall intelligence in all factors.

\section{Discussion and Conclusions}

The goal of our research was to better understand how to increase perceived robot intelligence through selection of ambient adapted voices, specifically using non-loudness features, in opposition to previous literature.

The first contribution was developing a protocol for collecting ambiance-modified voices through Zoom to mitigate the difficulties with in-the-wild collection of vocal features in noisy ambiances. We found that we were able to replicate our desired ambiances to a reasonable degree through changes in lighting, sound, and images, as confirmed through both our validation study and the ambiance ratings in our perception study. These results are especially exciting as they suggest that simple, low-cost virtual simulations of ambient environments could be a suitable means for data collection toward usage on real robots. Further studies should be conducted to compare results of simulated versus in-the-wild contexts.

Next, we proposed clustering the voice features to extract primary voice styles, as opposed to a brute-force method of creating a separate voice for each ambiance. This method requires less data and computational intensity while being more transparent and interpretable. Testing robot voices selected based off these clusters, we found that the presence of a voice style in an ambiance from the Zoom study (Fig. \ref{barplot}) tended to match the preference of the matching voice style in that same ambiance in the on-robot study (Fig. \ref{1st}). This provides evidence supporting our overall procedure.

Thirdly, our results indicate that the Neutral voice (default on Pepper) was significantly NOT preferred in loud, high energy environments (Table II). Instead, a higher pitched voice with high pitch range and speed was preferred. Surprisingly, given both the literature on human vocal adaption and our human voice analysis results, we did not see our curated ``Lombard" voice outperforming Pepper's base Joyful voice in these loud environments. However, the designed ``Lombard" voice tended to have a higher comfort or social appropriateness ratings in the loud environments (loud bar, lively restaurant, nightclub) compared to the Neutral voice. One participant commented that the ``Lombard" voice was ``Robotic sounding but louder" in the nightclub setting. Since the robot volumes remained constant, this could suggest that the ``Lombard" voice with increased pitch could indeed increase perceived loudness. Finally, it should be noted that our in-person study did \emph{not} include any negative sentiment sentences. It remains to be tested whether the ``Lombard" voice could outperform the Joyful Pepper voice in these contexts (e.g. ``Sorry, we don't have that today."). As expected, the Neutral voice was found to be more socially appropriate and ambiance-aware in the quiet fine dining ambiance.


 Overall, other than the café and quiet bar, including those where there was no significance in the selection of preferred server, the preferred voice saw an significant increase in social appropriateness and ambiance awareness. This suggests that carefully selecting the correct voice for an ambiance can improve human perception of socially intelligent robots. Additionally, Fig. 10 depicts strong correlations between an increase in socially appropriate voice and competency.

It is important to note the limitations of the aforementioned study. First, although we collected clear audio data for a variety of different virtual ambient environments, we could not control for how one would adapt their voice virtually versus in a real ambient environment. Second, we could not control for group dynamics in the user study. More specifically, how one would interact, perceive, or interpret Pepper in isolation may be different to when not in isolation. Finally, we were limited to Pepper's vocal features and thus our work was tailored to that of Pepper. Introducing a variety of social robots in the future would allow us to observe if embodiment impacts the results.

Future work includes investigating the number of participants needed for the design process, other robots, and the effect of gender, age or culture on the results.
Real-time systems able to adapt a voice using a sensing-production loop is also an exciting area for future study; for instance, the robot may assess the distribution of frequencies in the ambiance and adapt its voice to avoid the frequency ranges already filled by background noise. Furthermore, the use of virtual reality could be explored to overcome the limitations in simulating the closed/open characteristics of space, which were not well differentiated in our perception study.

\section*{APPENDIX} \label{battery}
\footnotesize
Battery for speaker validation study:
\begin{itemize}
\footnotesize 
    \item This person's voice is socially appropriate for the scene.
    \item This person knows how they should present themselves in this ambient context.
    \item This person is aware of the surrounding ambiance. 
    \item This voice sounds like it was originally recorded in this ambiance. 
    \item This person knows how to adapt their voice this ambiance. 
    \item This person makes me feel comfortable. 
    \item I feel  uneasy listening to this person. 
    \item I feel at easy when conversing with this person. 
    \item It took a great deal of effort to understand what this person was saying.
    \item Certain words were difficult to understand.
    \item This person's speech was very clear. 
\end{itemize}



\bibliographystyle{IEEEtran}
\bibliography{IEEEabrv,IEEEexample.bib}

\end{document}